\documentclass[a4paper,11pt]{article}

\usepackage{fourier}
\usepackage{color}
 \usepackage{graphicx}
\usepackage{url}
\usepackage[affil-it]{authblk}
\usepackage{amsmath}
\usepackage{wrapfig}
\usepackage{gensymb}
\usepackage{tabularx}
\usepackage{array}
\usepackage{multirow}
\usepackage{wrapfig, blindtext}

\usepackage[T1]{fontenc}
\usepackage{times}
\usepackage{etoolbox,lipsum}

\begin{document}
\renewcommand{\thefootnote}{\alph{footnote}}

\title{Deep Multi-Task Networks For Occluded Pedestrian Pose Estimation}

%\author{Anonymous Submission}
\author{Arindam Das$^{1,2}$, Sudip Das$^3$, Ganesh Sistu$^4$, Jonathan Horgan$^4$, \\ Ujjwal Bhattacharya$^3$, Edward Jones$^2$, Martin Glavin$^2$, and Ciar\'{a}n Eising$^{2,5}$}

%\affil{Anonymous Affiliation}
\affil{$^1$Detection Vision Systems, Valeo India, $^2$National University of Ireland, Galway,\\ $^3$Indian Statistical Institute, Kolkata, India, $^4$Valeo Vision Systems, Ireland, \\ $^5$University of Limerick, Ireland}

\vspace{-2cm}

\date{}
\maketitle

\vspace{-0.5cm}

\begin{abstract}
%Pedestrian pose estimation becomes increasingly challenging when (i) parts of the human figures are occluded by some obstacles and (ii) in outdoor scene with crowded pedestrians.
Most of the existing works on pedestrian pose estimation do not consider estimating the pose of an occluded pedestrian, as the annotations of the occluded parts are not available in relevant automotive datasets. %This limitation makes a major bottleneck to make progress in this area of research.
For example, CityPersons, a well-known dataset for pedestrian detection in automotive scenes does not provide pose annotations, whereas MS-COCO, a  non-automotive dataset, contains human pose estimation. In this work, we propose a multi-task framework to extract pedestrian features through detection and instance segmentation tasks performed separately on these two distributions. Thereafter, an encoder learns pose specific features using %a domain invariant way 
an unsupervised instance-level domain adaptation method for the pedestrian instances from both distributions. The proposed framework has %been extensively studied on two benchmark datasets and it has
improved state-of-the-art performances of pose estimation, pedestrian detection, and instance segmentation. %in scenarios such as heavy occlusions (HO) and reasonable + heavy occlusions (R+ HO).
\end{abstract}


\vspace{-0.3cm}

%%%%%%%%%%%%%%%%%%%%%%

\section{Introduction}
%Pedestrian pose estimation is an essential and critical task in many visual perception based intelligent systems such as driver assistance setup \cite{horgan2015vision} for Autonomous Vehicles (AVs). Although pose estimation has been an active area of research since early 2000s \cite{aggarwal1999human}, however the progression has been significant over the past few years due to the development of advanced deep learning methods \cite{insafutdinov2016deepercut, kocabas2018multiposenet, sun2019deep, das2020end} for the same and availability of large-scale datasets \cite{andriluka20142d, lin2014microsoft, li2019crowdpose}. 

Human Pose Estimation (HPE)
is a widely deployed computer vision application \cite{chen2020monocular}, along with semantic segmentation \cite{segdasvisapp19}, and object tracking and trajectory prediction \cite{sridevi2021object}. % and soiling detection \cite{das2019soildnet}. % and shadow removal \cite{dasgupta2022unshadownet}.
Addressing HPE for occluded pedestrians was first reported in ClueNet \cite{kishore2019cluenet} and later \textit{person-to-person} occlusion was investigated in \cite{das2020end}. In this paper, we propose a novel \textit{end-to-end two stage fully convolutional network} for the purpose of estimating pose of occluded pedestrians, in the context of a Multi-Task Learning (MTL) architecture where object detection and segmentation are performed. The proposed framework can 1) learn pedestrian pose estimation (PPE) from a dataset that does not contain pose annotations and 2) accurately estimate the pose of the occluded parts of the human body without the annotations of the same occluded parts. To address the issue of the non-availability of pose annotations during training in the CityPersons \cite{zhang2017citypersons} dataset, we consider a related dataset (MS-COCO \cite{lin2014microsoft}) to get adequate supervision specific to PPE during training. To achieve this, we apply unsupervised instance level domain adaptation on each pedestrian into the target domain. Here, the dataset with human pose annotations acts as the source domain, whereas the samples with only pedestrian detection and segmentation annotations become the target domain in domain adaptation. The main contributions of this study are: 1) the proposal of a two-stage end-to-end fully convolutional network to perform occluded PPE, 2) preserving the information of detection and segmentation in an MTL architecture, and 3) in achieving state-of-the-art performance on pedestrian detection, instance segmentation and PPE tasks respectively. 

% \begin{enumerate}
%     \itemsep0em 
%     \item An \textit{end-to-end multi-task fully convolutional network} to perform occluded pedestrian pose estimation
%     \item We propose Spatio-Semantic Consistency Loss .
%     \item We achieved state-of-the-art results on .
% \end{enumerate}

%\input{include/related_work.tex}

\section{Methodology}
\begin{figure*}[]
    \centering
    \includegraphics[width=0.9\textwidth]{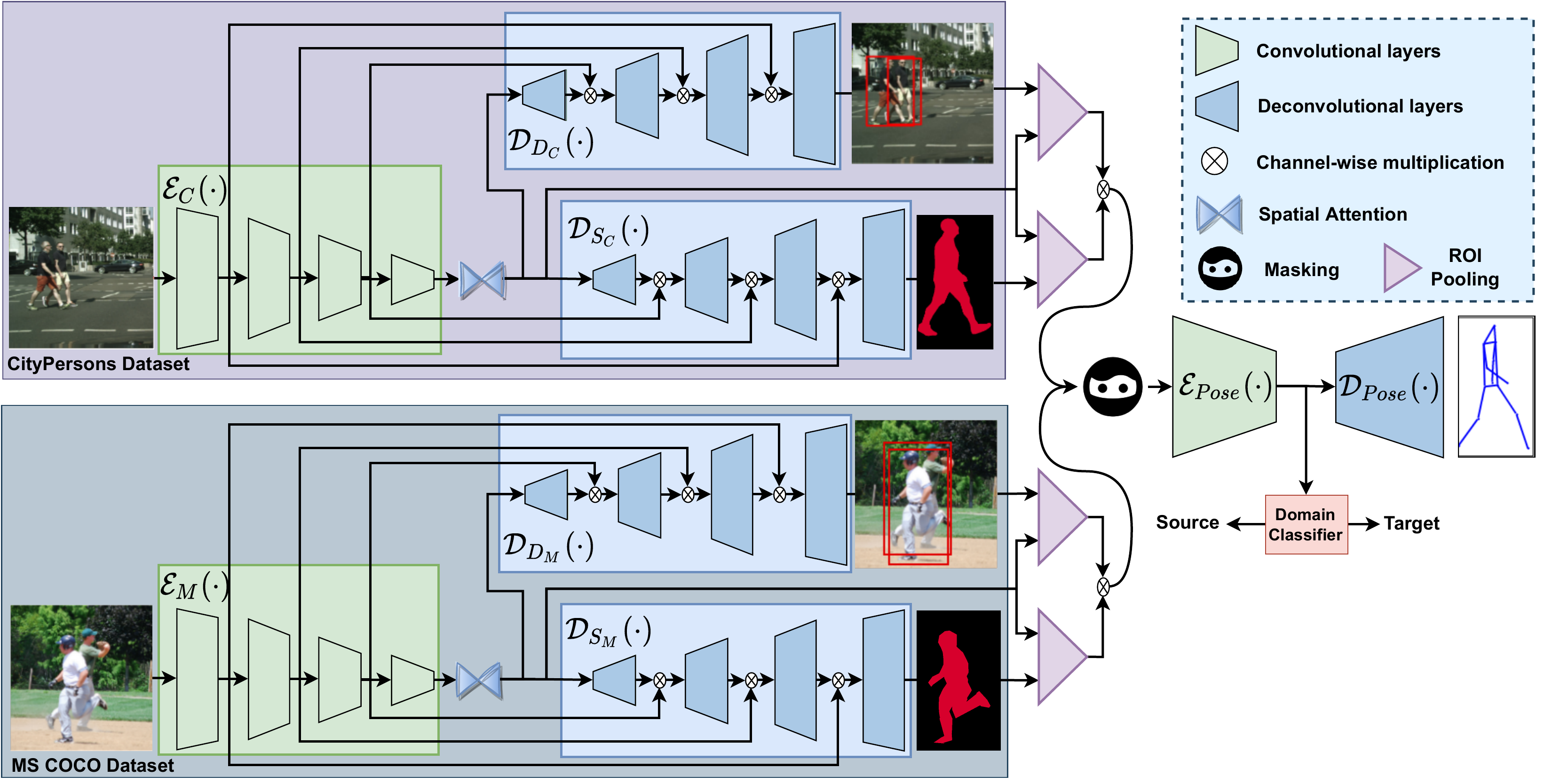}
    \vspace{-4mm}
    \caption{
    \textbf{Our proposed MTL architecture for end-to-end PPE.}
    }
    \label{fig:proposed_pose}
    \vspace{-0.5cm}
\end{figure*}

\subsection{Distribution Specific Multi-task Learning}
Two distinct multi-task learning (MTL) networks are setup for two different data distributions - CityPersons and MS COCO. Each MTL network aims to extract domain specific features of human instances to perform human detection and instance segmentation respectively. Our proposed framework is illustrated in Figure \ref{fig:proposed_pose}. Feature Pyramid Network (FPN) style encoder is used in each distribution specific MTL instance. We denote these as $\mathcal{E}_{C}(\cdot)$ and $\mathcal{E}_{M}(\cdot)$ respectively. We use a set of deconvolution layers to perform instance segmentation ($\mathcal{D}_{S_C}(\cdot)$ and $\mathcal{D}_{S_M}(\cdot)$) and detection  ($\mathcal{D}_{D_C}(\cdot)$ and $\mathcal{D}_{D_M}(\cdot)$) in two instances of MTL. Two independent MTLs are necessary here as the distribution of the datasets considered in this work is not close. One of the main advantages of having distribution specific MTL is to exploit these two domains independently to generate strong distribution specific features. We use spatial attention in between encoder and task specific decoders. This basically applies global average pooling on feature maps to generate vector and that is multiplied with the same feature map to generate most attentive area from the features. This technique helps to preserve the spatial consistency in feature space.

\subsection{Distribution Invariant PPE}
As the pedestrian dataset does not provide pose annotations, we apply instance level domain adaptation from another dataset that contains the details of human pose. Pedestrian instances from the detection and segmentation decoder are projected back to the last layer of encoded feature maps using ROI pooling, that are masked and fed as input to the encoder ($\mathcal{E}_{Pose}(\cdot)$) for PPE. The motivation to apply \textit{instance-level} domain adaption is to minimize the domain shift between two different distributions of the same category.  %Unsupervised adversarial technique is followed here to reduce the gap in distributions.
In this unsupervised domain adaptation setup, when a sample from source domain passes through the pose encoder, then the weights are updated in $\mathcal{E}_{Pose}(\cdot)$, domain classifier ($\mathcal{D}_{C}(\cdot)$) and $\mathcal{D}_{Pose}(\cdot)$ but $\mathcal{D}_{Pose}(\cdot)$ is not updated for the input from target domain. This means the features extracted by $\mathcal{E}_{Pose}(\cdot)$ are fed to $\mathcal{D}_{C}(\cdot)$ to determine the actual source distribution (i.e., CityPersons or MS COCO). In this adversarial learning setup, $\mathcal{D}_{C}(\cdot)$ promoted the accurate classification of the input domain, while $\mathcal{E}_{Pose}(\cdot)$ encourages the generation of better domain invariant features specific to human instances.
%$\mathcal{D}_{C}(\cdot)$ was trained using the human instances from both datasets. 

To train the proposed framework in end-to-end fashion, we express overall MTL loss function, $\mathcal{L}_{total}$ as the simple weighted combination of other losses from the detectors ($\mathcal{L}_{det_c}$, $\mathcal{L}_{det_m}$), instance segmentation ($\mathcal{L}_{seg_c}$, $\mathcal{L}_{seg_m}$), domain classification ($\mathcal{L}_{dc}$) and PPE ($\mathcal{L}_{pe}$) as,
\begin{equation}
    \mathcal{L}_{total} = \mathcal{L}_{det_c} + \mathcal{L}_{det_m} +  \alpha\mathcal{L}_{seg_c} + \alpha\mathcal{L}_{seg_m} +  \beta\mathcal{L}_{dc} + \gamma\mathcal{L}_{pe}
\end{equation}
where $\mathcal{L}_{det_c}$, $\mathcal{L}_{seg_c}$ are the losses of detection and segmentation tasks trained on CityPersons dataset. Likewise, losses $\mathcal{L}_{det_m}$ and $\mathcal{L}_{seg_m}$ are obtained from an MTL trained on MS COCO. $\alpha$, $\beta$ and $\gamma$ are the weights corresponding to instance segmentation, domain classifier and PPE losses respectively. For the instance segmentation task, we use binary cross-entropy loss. For object detection task, we adopt the loss from \cite{dasgupta2022spatio}. We follow  the loss function for PPE and $\mathcal{L}_{dc}$ as described in \cite{kishore2019cluenet} and \cite{kocabas2018multiposenet}.

\section{Experimentation Details}

\subsection{Datasets and Implementation Details}

The proposed method is trained and evaluated on the two publicly available datasets already mentioned - CityPersons and MS COCO. Annotations of detection and instance segmentation specific to human instances are used from both datasets, along with pose information from MS COCO. Out of $17$ available key points, we use $13$ (as discussed in \cite{kishore2019cluenet}). %Only human instances were considered from both datasets. 
Two MTL networks specific to detection and instance segmentation tasks were pre-trained on CityPersons and MS COCO before initiating the second stage of training for PPE. Curriculum Learning for Mask and Predict strategy was used to gradually increase the masking percentage as the training progresses. Momentum Optimizer with $0.9$ and an initial learning rate of $0.01$ was used. After each $15k$ iterations, the learning rate is reduced by a factor of $10$. The weights $\alpha$, $\beta$ and $\gamma$ are set to $0.5$, $1$, $1$ respectively. Data augmentation methods, such as applying random flipping, blurring, brightness, etc., are added to make the proposed framework more robust. Training was completed on two Nvidia Tesla P$6$ GPUs with batch size set to $1$.  

%CityPersons dataset contains the annotations for pedestrian detection and segmentation tasks. Training, validation, and testing samples distribution of this dataset is $2695$, $500$ and $1575$ respectively. We use key points and segmentation mask from Microsoft COCO dataset to train our network. Out of 17 available key points, we use 13 as considered in \cite{kishore2019cluenet}.

%\paragraph{Evaluation Metrics:}
%To evaluate our proposal, we use log average Miss Rate (MR) to report the error on the detection performance of pedestrians. In our experiments, different occlusion scenarios such as Reasonable (R) with [.65, inf] visibility, Heavy Occlusion (HO) with [.20, .65] visibility and Reasonable+Heavy occlusion (R + HO) with [.20, inf] visibility are considered to perform PPE. Standard IoU and Average Precision (AP) are used to quantify the segmentation and PPE results.

\subsection{Results}
\begin{wrapfigure}{r}{0.5\linewidth}
    \vspace{-3mm}
    \centering
    \includegraphics[width=0.12\textwidth, height=0.237\textwidth]{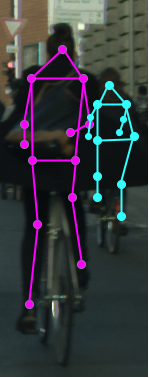}
    \includegraphics[width=0.15\textwidth, height=0.238\textwidth]{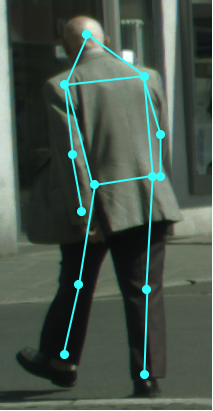}
    \includegraphics[width=0.15\textwidth]{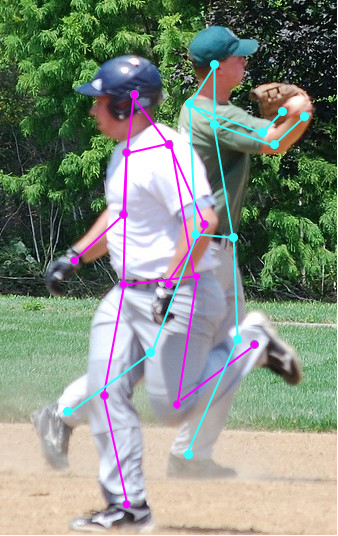}
    \vspace{-3mm}
    \caption{
    \textbf{Sample outputs of the pose estimation.}
    }
    \label{fig:samples}
    \vspace{-3mm}
\end{wrapfigure}

Table \ref{tab:pose_estimation_results} shows the performance of PPE of the proposed model and compares with recently published methods. Since the annotations of the occluded parts of the pedestrians are not available, we created such occluded pedestrians by masking (with different percentages) random parts of fully visible pedestrians \cite{das2020end} and compared with existing techniques as presented in Table \ref{tab:pose_estimation_occlusion_results}. The proposed approach has clearly improved the existing benchmark in estimating pose for the occluded pedestrians. Comparison study using Miss Rate (MR) on the  validation set of CityPersons for pedestrian detection task with a few existing methods is shown in Table \ref{tab:detection_results} where a few specific scenarios such as Reasonable (R) with [.65, inf] visibility, Heavy Occlusion (HO) with [.20, .65] visibility and Reasonable+Heavy occlusion (R + HO) with [.20, inf] visibility are considered. Fig. \ref{fig:samples} shows the qualitative results on CityPersons and MS COCO dataset for PPE of the proposed model. Table \ref{tab:instance_segmentation_results} provides the results of instance segmentation of the proposed framework that slightly improves the SOTA performance for both categories - person and rider. As part of ablation study, different standard backbone encoders are tested using average precision metric for PPE and among these ResNeXt-101 outperformed other encoders as presented in Table \ref{tab:different_encoder_results}.

%%%%%%%%%%%%%%%%%%%%%%%%%%%%%%%%%%%%%%%%%%%%%%%%%%%%%%%%%%%

\begin{table}
\parbox{.4\linewidth}{
\centering
\caption{\small Comparison using AP for PPE task on MS COCO dataset.}

\scalebox{.6}{
\begin{tabular}{|c|c|c|c|c|c|}
\hline
\textbf{Model} & $\textbf{AP}$ & $\mathbf{AP_{50}}$ & $\mathbf{AP_{75}}$ & $\mathbf{AP_{M}}$ & $\mathbf{AP_{L}}$ \\ \hline \hline 

MultiPoseNet \cite{kocabas2018multiposenet}      &  69.6  &   86.3     &  76.6  & 65.0    &  76.3     \\ \hline
% Megvii \cite{chen2018cascaded}      &  73.0  &  \textbf{91.7}  &  \textbf{80.9}  & 69.5   &  78.1      \\ \hline
CFN  \cite{huang2017coarse}      &  72.6  &   86.7     &  69.7  & 78.3   &  79.0      \\ \hline
 ClueNet\cite{kishore2019cluenet}    &    73.9 &  89.6   &  \textbf{78.2}       &  70.9  & 79.1 \\ \hline
 \cite{das2020end}    &  74.2 &  89.9  & 74.9  &  79.3  & 76.6 \\ \hline
 \textbf{Ours}    &  \textbf{75.7} &  \textbf{90.3}  & 76.3  &  \textbf{80.7}  & \textbf{79.5} \\ \hline
\end{tabular}}
\label{tab:pose_estimation_results}
}
\hfill
\parbox{.55\linewidth}{

\centering
\caption{\small Comparative study of PPE with SOTA models having different occlusion percentages on MS-COCO dataset.}

\scalebox{.7}{
\begin{tabular}{|c|c|c|c|c|c|c|}
\hline
\multirow{2}{*}{\textbf{Model}} & \multicolumn{6}{c|}{Occlusion \%} \\
  & \multicolumn{1}{c}{\textbf{20\%}} & \multicolumn{1}{c}{\textbf{30\%}} & \multicolumn{1}{c}{\textbf{40\%}} & \multicolumn{1}{c}{\textbf{50\%}} & \multicolumn{1}{c}{\textbf{60\%}} & \multicolumn{1}{c|}{\textbf{70\%}} \\ \hline \hline
ClueNet \cite{kishore2019cluenet} & 88.06 & 83.93 & 79.8 & 73.4 & 64.0 & 58.8 \\ \hline
\cite{das2020end} & 90.3 & 84.31 & 81.2 & 74.06 & 64.9 & 59.1 \\ \hline
\textbf{Ours} & \textbf{92.0} & \textbf{85.9} & \textbf{82.4} & \textbf{75.3} & \textbf{65.7} & \textbf{59.3} \\ \hline
\end{tabular}}
\label{tab:pose_estimation_occlusion_results}

} \vspace{-0.3cm}
\end{table}

%%%%%%%%%%%%%%%%%%%%%%%%%%%%%%%%%%%%%%%%%%%%%%%%%%%%%%%%%%%

\begin{table}
\parbox{.45\linewidth}{
\centering
\caption{\small MR based comparison of SOTA models for pedestrian detection on  CityPersons.}
\scalebox{.6}{
\begin{tabular}{|c|c|c|c|}
\hline
\textbf{Model} & \textbf{R} & \textbf{HO} & \textbf{R+HO} \\ \hline \hline
  Faster RCNN   & 15.52  &  64.83  &  41.45    \\ \hline
  Tao et al. \cite{song2018small}  &  14.4  & 52.0   &  34.24   \\ \hline
  OR-CNN \cite{zhang2018occlusion}  &  \textbf{11.0}  & 51.0   &  36.11   \\ \hline
  ClueNet \cite{kishore2019cluenet} & 11.87  & 47.68  &  30.84   \\ \hline
  \cite{das2020end} & 13.29  & 46.07  &  29.13   \\ \hline
  \textbf{Ours} & 12.01 & \textbf{44.7}  &  \textbf{27.8}   \\ \hline
\end{tabular}}
\label{tab:detection_results}
}
\hfill
\parbox{.5\linewidth}{
\caption{\small Comparison using IoU for instance segmentation on CityPersons.}
\label{tab:instance_segmentation_results}
\centering
\scalebox{.7}{
\begin{tabular}{|c|c|c|c|} 
\cline{1-4}
\textbf{Model}  & \textbf{Training Data}  & \textbf{Person} & \textbf{Rider}   \\
\hline \hline
Mask-RCNN  & CityPersons + COCO & 34.8 & 27.0\\ 
\cline{1-4}
% Davy et al.  \cite{neven2019instance} & CityPersons & 34.5 & 26.1  \\ 
% \cline{1-4}
PANet  \cite{liu2018path} & CityPersons + COCO  & 41.5 & 33.6 \\ 
\cline{1-4}
\cite{das2020end}  & CityPersons + COCO & 42.1 & 33.9\\
\cline{1-4}
\textbf{Ours}  & CityPersons + COCO & \textbf{42.7} & \textbf{34.7}\\
\cline{1-4}
\end{tabular}
}}
 \vspace{-0.3cm}
\end{table}

%%%%%%%%%%%%%%%%%%%%%%%%%%%%%%%%%%%%%%%%%%%%%%%%%%%%%%%%%%%

\begin{table}[!h]
\centering
\caption{\small Ablation study on different backbone encoders for PPE on test data of MS-COCO.}
\label{tab:different_encoder_results}
\vspace{1mm}
\scalebox{.8}{
\begin{tabular}{|c|c|c|c|c|c|}
\hline
\textbf{Backbone} & $\textbf{AP}$ & $\mathbf{AP_{50}}$ & $\mathbf{AP_{75}}$ & $\mathbf{AP_{M}}$ & $\mathbf{AP_{L}}$ \\ \hline \hline 
%VGG-19  &  63.9  &   80.7     &  70.6  & 58.0   &  70.3      \\ \hline
ResNet-50  &  69.7  &   86.3     &  71.94  & 64.2   &  71.1      \\ \hline
ResNet-101 &  71.4  &   87.8     &  72.1  & 76.1    &  74.3     \\ \hline
ResNeXt-101 &  \textbf{75.7} &  \textbf{90.3}  & \textbf{76.3}  &  \textbf{80.7}  & \textbf{79.5}      \\ \hline
\end{tabular}}
\vspace{-0.3cm}
\end{table}

\section{Conclusion}
In this work, an end-to-end two stage network is developed that is trained in an unsupervised manner to accurately estimate the pose of pedestrians regardless the level of occlusion. We apply unsupervised domain adaptation at instance level to reduce the distribution gap between two set of features obtained from two distinct MTL setup. Experimental results demonstrate the robustness of the proposed strategy and provide strong confirmation as it improved respective state-of-the-art results on PPE, pedestrian detection, and instance segmentation.

%\input{include/acknowledgement.tex}

%\bibliographystyle{apalike}
%\bibliography{imvip}

\begingroup
\let\section\subsubsection
\makeatletter
\renewcommand\@openbib@code{\itemsep\z@}
\makeatother
\bibliographystyle{apalike}
\bibliography{imvip}
\endgroup

\end{document}